\documentclass{article}

\usepackage{arxiv}

\usepackage[utf8]{inputenc} 
\usepackage[T1]{fontenc}    
\usepackage{hyperref}       
\usepackage{url}            
\usepackage{booktabs}       
\usepackage{amsfonts}       
\usepackage{nicefrac}       
\usepackage{microtype}      
\usepackage{lipsum}		
\usepackage{graphicx}
\usepackage{natbib}
\usepackage{doi}

\usepackage{graphicx} 
\usepackage{subcaption} 
\usepackage{amsmath,amsfonts,amsthm,bm}
\usepackage{xcolor} 
\usepackage[toc,page]{appendix}
\usepackage{bbm}
\usepackage{txfonts}
\usepackage{booktabs} 
\usepackage{multirow} 
\usepackage{comment}
\usepackage{arydshln} 
\usepackage{natbib} 
\usepackage{adjustbox}
\usepackage{empheq} 
\usepackage{makecell}
\usepackage{tcolorbox} 
\usepackage{pifont} 

\usepackage{algorithm}
\usepackage{algpseudocode}
\floatname{algorithm}{Pseudocode}

\theoremstyle{definition}

\usepackage{setspace}
\usepackage{hyperref} 
\hypersetup{
    colorlinks=true,                
    breaklinks=true,                
    linkcolor= blue,                
    bookmarksopen=false,
    citecolor=blue,
    linkbordercolor=blue
}
\title{Decision-centric fairness: \\Evaluation and optimization for resource allocation problems}

\author{ 
    Simon De Vos
\thanks{Corresponding author: \texttt{simon.devos@kuleuven.be}} 
\\
	KU Leuven\\
	\And
    Jente Van Belle\\
	KU Leuven\\
	\And
    Andres Algaba \\
	Vrije Universiteit Brussel \\
        \And
    Wouter Verbeke \\
	KU Leuven \\
        \And
    Sam Verboven \\
        Vrije Universiteit Brussel\\
}

\renewcommand{\shorttitle}{Decision-centric fairness: Evaluation and optimization for resource allocation problems}

\begin{document}
\maketitle

\begin{abstract}
Data-driven decision support tools play an increasingly central role in decision-making across various domains. In this work, we focus on binary classification models for predicting positive-outcome scores and deciding on resource allocation, e.g., credit scores for granting loans or churn propensity scores for targeting customers with a retention campaign. 
Such models may exhibit discriminatory behavior toward specific demographic groups through their predicted scores, potentially leading to unfair resource allocation. 
We focus on demographic parity as a fairness metric to compare the proportions of instances that are selected based on their positive outcome scores across groups.
In this work, we propose a decision-centric fairness methodology that induces fairness only within the decision-making region---the range of relevant decision thresholds on the score that may be used to decide on resource allocation---as an alternative to a global fairness approach that seeks to enforce parity across the entire score distribution.
By restricting the induction of fairness to the decision-making region, the proposed decision-centric approach avoids imposing overly restrictive constraints on the model, which may unnecessarily degrade the quality of the predicted scores.
We empirically compare our approach to a global fairness approach on multiple (semi-synthetic) datasets to identify scenarios in which focusing on fairness where it truly matters, i.e., decision-centric fairness, proves beneficial.
\end{abstract}


\section{Introduction}\label{SEC:INTRODUCTION}

In an increasingly data-driven world, algorithms and machine learning models play a crucial role in business decision-making. 
In this paper, we focus on predictive models that are used to optimize resource allocation, particularly on binary classification models that predict positive-outcome scores for this purpose.
Such models are widely used across various domains, including marketing, where retention incentives are offered based on churn propensity scores \citep{verbeke2011churn,verbeke2012churn};
credit risk management, where loans are granted based on default risk assessments \citep{lessmann2015credit};
and fraud detection, where investigative resources are allocated based on predicted fraud risk
\citep{vanlasselaer2017fraud}.
These models can exhibit discriminatory behavior toward specific demographic groups through their predicted scores---that is, the predicted scores may follow different distributions across demographic groups---potentially leading to unfair resource allocations.
Such discriminatory behavior can originate from biases in the historical data that is used to train the models or from inherent differences between groups in their tendency to belong to the positive class, which---although statistically justified---may be considered unacceptable discrimination when acted upon.

Fairness is central to the acceptability of algorithm-informed decisions, particularly in domains where these decisions can significantly affect individuals' access to resources or opportunities \citep{hardt2016}. While much of the existing research has focused on preventing gender-based discrimination in pricing to comply with regulatory standards \citep{lindholm2022discrimination,frees2023discriminating,xin2024antidiscrimination}, the relevance of fairness extends well beyond pricing models to key business functions such as credit risk assessment, targeted marketing, and fraud detection. Fairness is closely tied to principles embedded in non-discrimination laws, particularly in the EU and US, which emphasize equitable outcomes across demographic groups and provide a foundation for fairness criteria like demographic parity \citep{EU2012Guidelines}. Similar to the pricing context, discrimination in credit risk management \citep{hurlin2024fairness, garcia2024algorithmic} and fraud detection is legally prohibited (e.g., to prevent gender-based discrimination and ethnic profiling). Additionally, in marketing, fairness considerations are seen by some companies as crucial for building and maintaining both a diverse customer base and a positive reputation.

\begin{figure}[tb]
    \centering
    \begin{subfigure}[b]{0.48\textwidth}
        \centering
        \includegraphics[height=4.5cm]{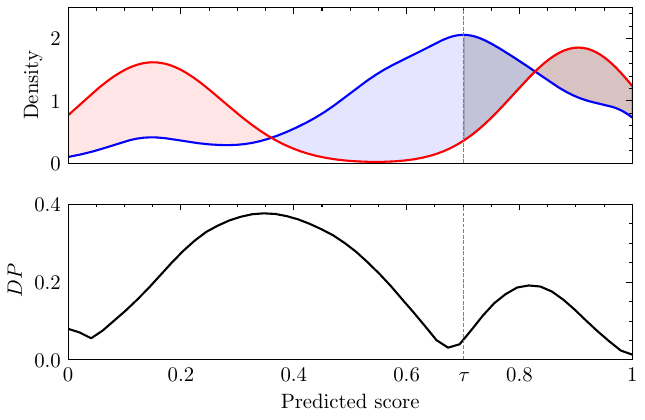}
        \caption{No fairness induction}
    \end{subfigure}
%
%
    \begin{subfigure}[b]{0.48\textwidth}
        \centering
        \includegraphics[height=4.5cm]{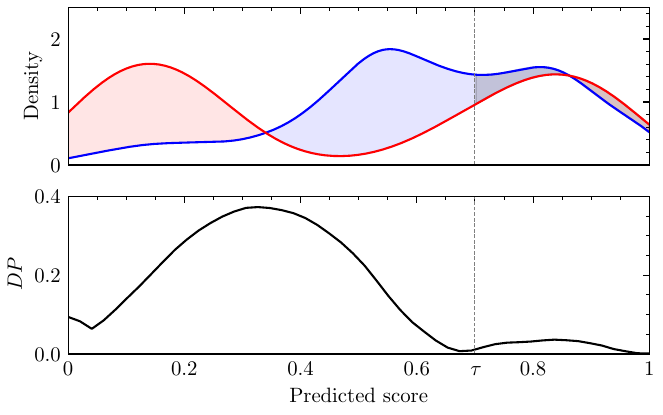}
        \caption{Induced decision-centric fairness}
    \end{subfigure}
    \caption{
    Densities of predicted scores $\tilde{y}$ for two demographic groups (with protected attributes $s = 0$ and $s = 1$, in blue and red, respectively), along with the corresponding demographic parity ($DP$) across all possible thresholds.
    By inducing decision-centric fairness, we aim to achieve demographic parity in the decision-making region, i.e., where $\tilde{y} > \tau$, to ensure a proportionally equal number of positive outcomes across the two groups at all thresholds within this region.}
    \label{fig:intro_toy}
\end{figure}

Traditional approaches to evaluating algorithmic fairness typically rely on output-based metrics that assess disparities in average predictions or error rates between demographic groups \citep{makhlouf2021}.
These approaches are built on the idea that protected attributes, such as gender or race, should not impact a predicted score (e.g., a customer's default risk score).
Although such metrics provide a general view, they often overlook more subtle nuances in algorithmic behavior \citep{khan2023unbearable}. Recent work has expanded fairness analyses by incorporating higher-order moments of the output distribution, such as the variance \citep{khan2023fairness}, and by comparing entire output distributions \citep{han2023}.
More specifically, \citet{han2023} propose distribution-level variants of demographic parity,
a fairness metric that compares the proportions of positive class predictions---the predictions on which we would potentially act in a resource allocation setting---across groups.
While \citet{han2023} focus on model evaluation from a fairness perspective,
\citet{peeperkorn2024fairness} show that these distribution-level fairness notions can be used to develop predictive models that are intrinsically more fair.
Building on these works, we propose a pragmatic \textit{decision-centric fairness approach} to classification for resource allocation optimization.
Specifically, 
rather than focusing on inducing demographic parity across the entire output distribution (i.e., ensuring a proportionally equal number of positive outcomes at all possible decision thresholds on the predicted scores),
which we term a \textit{global fairness approach},
we propose to concentrate on the "decision-making region" by inducing parity only within the range of relevant thresholds used for resource allocation, as visualized in Figure~\ref{fig:intro_toy}.
For example, in a customer retention campaign, interventions are typically targeted at customers with a high predicted churn propensity score. Since resource allocation decisions affect only instances within this high-risk-score region, fairness should be enforced within this region, across all relevant application-dependent thresholds.
The proposed decision-centric approach aims to ensure fairness where it matters, while achieving better predictions compared to a global fairness approach. The latter imposes overly strict constraints on the model to enforce fairness also outside the decision-making region, where it will not affect real-world decisions, potentially degrading the predictive quality of the generated scores more than necessary.

While achieving fairness in terms of demographic parity is straightforward using a group-dependent decision threshold post-hoc when decisions are made in batch,
optimizing classification models to be inherently more fair without focusing on a single decision threshold is useful in common online decision-making settings (i.e., on a continuous basis).
In such settings, resource constraints---such as marketing budgets, loan-granting capacity, or investigative resources---may change over time \citep{ali2014dynamic}, causing the decision threshold to vary within a certain decision-making region. 
In these cases, setting group-dependent decision thresholds post hoc to achieve parity is not possible, and retraining the model each time a new threshold (within the decision-making region) is adopted due to changing resource constraints can be (too) costly.

Our main contributions are as follows: (i) we introduce and formalize the concept of decision-centric fairness for resource allocation optimization; (ii) we propose a decision-centric fairness approach to optimize classification models that are used for resource allocation; (iii) we propose a decision-centric predictive performance metric for classification models;
and (iv) we empirically compare our proposed decision-centric fairness methodology to a global fairness approach on multiple (semi-synthetic) datasets to identify scenarios where, from a decision-centric evaluation perspective, focusing on fairness only where it truly matters, outperforms imposing fairness globally across the output domain.

The remainder of this paper is structured as follows. 
In Section~\ref{SEC:LITERATURE}, we formalize the problem setting and discuss common fairness metrics and related work. Our proposed decision-centric fairness optimization methodology is presented in Section~\ref{SEC:METHODOLOGY}.
Section~\ref{SEC:EXPERIMENTAL_DESIGN} outlines the experimental design, with results presented and discussed in Section~\ref{SEC:RESULTS+DISCUSSION}.
Finally, Section~\ref{SEC:CONCLUSION} presents our conclusions and outlines possible future research directions.

\section{Background and related work} \label{SEC:LITERATURE}
In this section, we formalize the problem setting, provide a discussion of common fairness metrics, and motivate the use of demographic parity in business decisions.

\subsection{Classification and resource allocation}
Suppose that we have a data set $\mathcal{D}=\{(\bm{x_i},y_i,s_i)\}_{i=1}^N$ with $N$ the number of instances with each a feature vector $\bm{x}_i \in \mathbb{R}^k$, a binary response variable $y_i \in \{0,1\}$, and a binary protected attribute $s_i \in \{0,1\}$. Let $f:\mathbb{R}^{k+1} \rightarrow [0,1]$ be a model that maps instances to a score $\tilde{y} \in [0,1]$ which allows ranking instances from low to high score for belonging to the positive class. A predicted class $\hat{y} \in \{0,1\}$ is obtained by setting a decision threshold $\tau \in [0,1]$.
Instances with a predicted score below the threshold, $\tilde{y} < \tau$, are assigned to the negative class, while those with a score above the threshold are assigned to the positive class.
In resource allocation applications, resources are allocated to instances that are classified in the positive class based on the predicted score and the adopted threshold.
In practice, the decision threshold is often determined post-training, as resource constraints that affect the threshold may only be known at runtime (e.g., the available investigative capacity for fraud detection), and/or because the threshold is to be optimized at runtime (e.g., to maximize the profitability of a retention campaign by taking into account the customer lifetime value of customers that are classified as churners \citep{verbeke2012churn}).

\subsection{Fairness notions} \label{SEC:FAIRNESS_NOTIONS}
We focus on situations in which algorithmic fairness issues may arise as a result of using model scores in combination with a (variable) decision threshold $\tau$ to decide whether or not resources are allocated \citep{weerts2022does}. Specifically, we consider settings where an action (e.g., offering a retention incentive) is taken for instances with a model score $\tilde{y} \geq \tau$ \citep{kozodoi2022fairness}. A standard approach to determine whether a classification model conforms to a given notion of fairness is to evaluate its outcome distribution with respect to a number of protected attributes, such as gender or race \citep{fazelpour2021algorithmic}. However, the current literature on algorithmic fairness offers a wide range of fairness notions and metrics \citep{barocas_fairml_2019,makhlouf2021}.

\paragraph{Direct vs.\ indirect discrimination} Algorithmic discrimination comes in two types \citep{barocas_fairml_2019}: direct discrimination, where decisions are based on protected attributes, and indirect discrimination, where decisions hurt protected groups even without explicitly using protected attributes \citep{prince_proxy_2019}. Direct discrimination is typically avoided by excluding protected attributes or perfectly correlated variables \citep{barocas_law_2016}. Therefore, this paper focuses on indirect discrimination. Even without using sensitive data directly, discriminatory behavior can occur through proxy features \citep{lindholm2024fair}. For example, a churn model might offer a retention incentive based on customer usage, but if that usage pattern is linked to a protected characteristic, the incentive may end up favoring one group over another \citep{adams_2023_proxy}.

\paragraph{Individual vs.\ group fairness} Fairness metrics are typically categorized into two primary types \citep{fazelpour2021algorithmic}: individual fairness and group fairness. Individual fairness is based on the principle that similarly situated individuals should receive comparable outcomes \citep{dwork_2012_fairness}, yet its practical application is hindered by the challenge of defining a robust, context-independent metric for similarity \citep{fleisher2021s}. Alternative formulations of individual fairness are provided by causal and counterfactual fairness measures, which rely on explicitly modeling hypothetical scenarios at the individual level \citep{kusner_2017_counterfactuals}. In contrast, group fairness evaluates statistical differences in outcomes across various demographic segments, facing the difficulty of selecting an appropriate metric—since even popular measures like demographic parity and equal opportunity are often mutually incompatible \citep{binns_apparent_2020,kleinberg_inherent_2016}. In this paper, we focus exclusively on group fairness, as it is more prevalent in practice and its enforcement and measurement are comparatively more straightforward \citep{kearns_2018_gerrymandering}.

\paragraph{Group fairness metrics} Group fairness evaluation methods typically translate philosophical or political fairness ideals into statistical parity metrics that apply to model outputs \citep{chouldechova2017fair}. For instance, demographic parity in classification compares the proportions of positive class predictions ($\hat{y}=1$) across groups. Alternatively, by incorporating ground truth labels ($y$), one can assess disparities in error rates—equality of opportunity, for example, compares true positive rates between groups \citep{pleiss2017fairness}. These conventional output-based metrics focus on first-order statistics, evaluating average outcomes or error rates across groups \citep{hardt2016}. Although these group fairness notions are widely used, many alternative definitions exist~\citep{makhlouf2021}. However, these output-based evaluations mainly rely on threshold-dependent first-order statistics and lack interpretability, potentially overlooking discriminatory behavior captured by higher-order moments or the broader prediction distribution~\citep{algaba2023lucid,mazijn2023lucid}.

\paragraph{Distribution-level fairness metrics}
Recent work expands on traditional group fairness metric—which typically focus solely on first moments—by incorporating higher-order statistics, such as the variance \citep{khan2023fairness, khan2023unbearable}, or by comparing entire output distributions \citep{han2023}. More specifically, \citet{han2023} argue that the standard demographic parity metric fails to detect unfairness due to its threshold dependency, i.e., a slight variation in the decision threshold may potentially re-introduce unfairness. Therefore, they propose two distribution-level variants of demographic parity: 
        \begin{itemize}
            \item Area Between Probability density function Curves (ABPC):
                \begin{equation}\label{eq:abpc}
                    \operatorname{ABPC}=\int_0^1\left|f_0(x)-f_1(x)\right| \mathrm{d} x,
                \end{equation}
                where $f_0(x)$ and $f_1(x)$ are the probability density functions (PDFs) of the predicted scores for two demographic groups, characterized by a protected attribute $s$, with $s=0$ and $s=1$, respectively.
            \item Area Between Cumulative density function Curves (ABCC):
                \begin{equation}\label{eq:abcc}
                    \mathrm{ABCC}=\int_0^1\left|F_0(x)-F_1(x)\right| \mathrm{d} x,
                \end{equation}
                where $F_0(x)$ and $F_1(x)$ are the cumulative distribution functions (CDFs) of the predicted scores for two demographic groups, characterized by a protected attribute $s$, with $s=0$ and $s=1$, respectively.
        \end{itemize}
    While we agree that threshold-sensitivity is a challenging problem when implementing demographic parity in practice, we argue in Section~\ref{SEC:METHODOLOGY} that the proposed ABPC and ABCC metrics are too rigid as they enforce demographic parity across the entire output distribution. This allows the use of decision thresholds across the entire output distribution, including regions where, in practice, the decision threshold would never be set due to resource constraints that limit the number of instances that can be acted upon (e.g., those that can be targeted with a retention campaign).
    
\subsection{Demographic parity in resource allocation} \label{SEC:DP_IN_RESOURCE_ALLOCATION}
Fairness in business operations is increasingly recognized as a fundamental concern
\citep{de2022algorithmic}. When allocating resources based on predictive models, businesses must ensure that their decision-making processes adhere to well-defined fairness criteria \citep{de2024explainable}. However, as fairness in algorithmic decision-making is inherently multifaceted, choosing an appropriate fairness notion is non-trivial \citep{corbett2023measure,mazijn_score_2021}. Many widely-used fairness metrics, such as demographic parity and equal opportunity, are mutually incompatible except under highly constrained conditions \citep{kleinberg_inherent_2016}. This incompatibility is further exacerbated by deep-rooted philosophical disagreements on which notions are most appropriate in different contexts \citep{binns_fairness_2018}. The choice between demographic parity and equal opportunity ultimately hinges on the evaluator's underlying assumptions and worldview \citep{friedler2016possibility}, making it imperative to ground fairness considerations in regulatory and ethical frameworks that guide real-world business decisions.

In the European Union, non-discrimination legislation and fairness requirements stemming from hard and soft law sources are highly influential in this context. For instance, the EU guidelines on discrimination in insurance explicitly require fairness in the access to and supply of goods and services \citep{EU2012Guidelines}. Similarly, ensuring fairness in insurance pricing and preventing discriminatory effects in customer engagement strategies are also legal obligations \citep{frees2023discriminating}. This broader interpretation of fairness aligns closely with the principle of demographic parity, which seeks to equalize the likelihood of favorable outcomes across demographic groups, independent of underlying differences in target variable distributions and potential historical injustices. Beyond EU law, the concept of demographic parity also intersects with US anti-discrimination regulations, particularly in the context of disparate impact analysis \citep{feldman2015certifying,radovanovic2022fairdea,zafar2019fairness}. Despite this apparent close connection between what is legally required and demographic parity, EU law does not foresee a specific implementation where fairness must be enforced across all potential model outputs \citep{hanson2025engineering}. In fact, if individuals fall outside the actionable range of scores regardless of whether a model has been globally or locally constrained, the legal impact remains unchanged when fairness is implemented locally. This suggests that constraints focused solely on the decision-making region could potentially fulfill legal fairness or non-discrimination objectives equally well as methods pursuing global demographic parity.

At the same time, pursuing global demographic parity often comes at the cost of predictive performance and calibration.
Enforcing demographic parity usually involves imposing constraints on the model that can reduce predictive performance \citep{dutta2020there,wick2019unlocking}. Furthermore, demographic parity and calibration are fundamentally incompatible unless the base rates for the positive class are identical between the demographic groups \citep{pleiss2017fairness}. These trade-offs are critical considerations for businesses that must balance fairness objectives with operational effectiveness and the reliability of predicted scores.

\section{Decision-centric demographic parity} \label{SEC:METHODOLOGY}
In this section, we first propose decision-centric variants of the distribution-level demographic parity metrics, ABPC and ABCC.
We then outline how these metrics can be leveraged to induce decision-centric fairness in classification models used for resource allocation.

\subsection{Evaluating decision-centric fairness} \label{SEC:METHODOLOGY_EVAL}

The two distribution-level variants of demographic parity proposed by \citet{han2023}, introduced in Section~\ref{SEC:FAIRNESS_NOTIONS}, are threshold-independent.
As discussed in Section~\ref{SEC:INTRODUCTION}, this is an important property, as predictive models are often deployed in online decision-making settings,
in which resource constraints may change over time \citep{ali2014dynamic},
causing the decision threshold to vary.
Additionally, the decision threshold may depend on instance-dependent costs and benefits \citep{verbeke2012churn}.
In a resource allocation context, however, we argue that these distribution-level variants of demographic parity,
which allow for the use of decision thresholds across the entire output distribution, are overly strict.
Specifically, they penalize deviations from demographic parity even in regions with predictions that will never be acted upon in practice due to the aforementioned resource constraints.
To address this, we propose decision-centric variants of the two distribution-level demographic parity metrics that only penalize unfairness within a relevant decision-making region:
\begin{equation}\label{eq:abpc_local}
    \mathrm{ABPC_{\tau}}=\int_{\tau}^1\left|f_0(x)-f_1(x)\right| \mathrm{d} x,
\end{equation}
\begin{equation}\label{eq:abcc_local}
    \mathrm{ABCC_{\tau}}=\int_{\tau}^1\left|F_0(x)-F_1(x)\right| \mathrm{d} x.
\end{equation}
That is, we adapt the previously defined ABPC (Equation~\eqref{eq:abpc}) and ABCC (Equation~\eqref{eq:abcc}) metrics by restricting their integration domains to the decision-making region $[\tau,1]$, making ABPC and ABCC special cases of the more general proposed ABPC$_{\tau}$ and ABCC$_{\tau}$ metrics.

\subsection{Inducing decision-centric fairness} \label{SEC:METHODOLOGY_TRAIN}

The ABPC and ABCC distribution-level demographic parity metrics \citep{han2023} allow us to \textit{evaluate} algorithmic fairness across full score distributions. 
Building on this work, \citet{peeperkorn2024fairness} operationalize a \textit{global fairness approach}, which allows \textit{inducing} algorithmic fairness across full score distributions, by training a neural classifier using a composite loss function that combines standard binary cross-entropy loss $\mathcal{L}_{\text{BCE}}$ with an unfairness penalty $\mathcal{L}_{\text{unfairness}}$ for deviations from demographic parity between the distributions of predicted scores for two groups defined by a protected attribute $s$:
    \begin{equation}
        \mathcal{L}=(1-\lambda) \cdot \mathcal{L}_{\text{BCE}}+\lambda \cdot \mathcal{L}_{\text{unfairness}},
    \end{equation}
where the hyperparameter $\lambda$ controls the trade-off between predictive performance and fairness. They propose to use Integral Probability Metrics (IPMs) \citep{shalit2017estimating}, notably the 1-Wasserstein distance\footnote{Efficient methods are available to approximate the distance and its gradients, enabling its use in a neural network objective function, as it can be directly minimized using common frameworks for neural network training.}, to quantify differences in predicted score distributions between demographic groups over the entire domain $[0,1]$. However, in resource allocation applications, fairness constraints applied globally (i.e., over the entire output domain) can impose unnecessary rigidity,
aiming to enforce fairness also outside the decision-making region where it will not affect real-world decisions,
potentially degrading the predictive quality of the generated scores more than necessary. Therefore, we propose a \textit{decision-centric fairness approach} specifically tailored to induce fairness only within the decision-making region (i.e., over the domain $[\tau, 1]$), thereby focusing only on the predictions that are relevant for resource allocation.

Our approach operationalizes this concept by applying the unfairness penalty exclusively to the top-$k\%$ of the predicted scores of each protected group:
\begin{equation}
    \mathcal{L}_{\text{unfairness}} = \text{IPM}(\tilde{y}_0^{(k\%)}, \tilde{y}_1^{(k\%)}),
\end{equation}
where $\tilde{y}_0^{(k\%)}$ and $\tilde{y}_1^{(k\%)}$ denote the distributions of the top-$k\%$ predicted scores for the protected groups with $s=0$ and $s=1$, respectively, and the IPM corresponds to the 1-Wasserstein distance~\footnote{An alternative, more directly aligned with the goal of inducing fairness in the decision-making region $[\tau, 1]$, is a quantile-based approach that characterizes the differences between the score distributions in a discretized manner using a histogram-based \citep{ustinova2016learning} unfairness penalty. Although this allows for the direct use of the decision threshold $\tau$---which defines the relevant decision-making region---in optimization, training proved unstable in preliminary experiments, as even very small values of $\lambda$ led to all predicted scores $\tilde{y}$ falling below $\tau$. The percentile-based approach prevents this downward biasing of predicted scores.}.
To determine the percentile threshold $k\%$, we first train a baseline unconstrained model (with $\lambda=0$) and set $k\%$ to match the proportion of instances with scores above the decision threshold $\tau$ on a validation set,
irrespective of the protected attribute $s$. In this way, the unfairness penalty specifically focuses on actionable instances.

\begin{figure}[t!]
    \centering
        \begin{subfigure}[t]{0.32\textwidth}
        \centering
        \includegraphics[height=3.5cm]{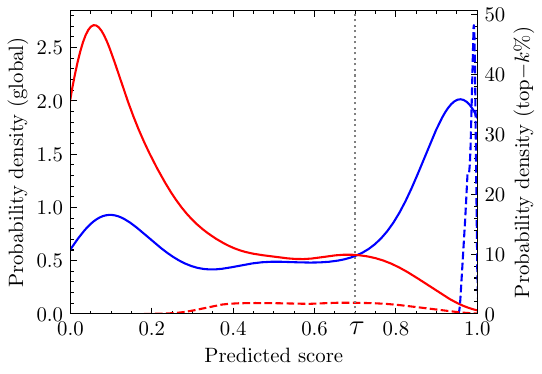}
        \caption{$\lambda=0.0$
        }
        \label{fig:DCF_optimization_0.0}
    \end{subfigure}
    \hfill
    \begin{subfigure}[t]{0.32\textwidth}
        \centering
        \includegraphics[height=3.5cm]{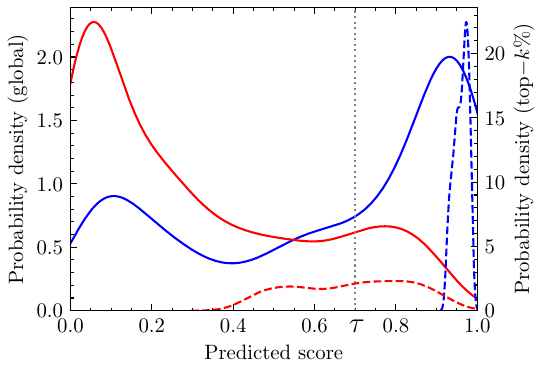}
        \caption{$\lambda=0.3$
        }
        \label{fig:DCF_optimization_0.3}
    \end{subfigure}
    \hfill
    \begin{subfigure}[t]{0.32\textwidth}
        \centering
        \includegraphics[height=3.5cm]{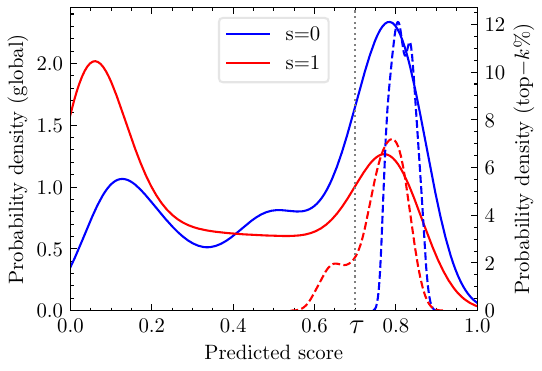}
        \caption{$\lambda=0.6$
        }
        \label{fig:DCF_optimization_0.6}
    \end{subfigure}
    \caption{The decision-centric fairness approach for different values of $\lambda$. Score distributions (PDFs) on a test set after training each model for 30 epochs are shown, split by the protected attribute ($s = 0$ and $s = 1$). 
    Solid lines represent the distributions of all predicted scores, while dotted lines represent the top-$k\%$ score distributions. The decision threshold $\tau$ is shown as a vertical line. An animated version of these plots is available in the \href{https://github.com/SimonDeVos/DCF}{GitHub repository} (\texttt{github.com/SimonDeVos/DCF}).}
    \label{fig:DCF_optimization}
\end{figure}

Conceptually, this means we alter the composition of the set of instances with $\tilde{y} \geq \tau$, as produced by a model without any penalty for unfairness, in order to induce greater fairness in the decision-centric demographic parity sense. This effect is illustrated in Figure~\ref{fig:DCF_optimization}, which shows how different values of the hyperparameter $\lambda$ impact the predicted score distributions for two demographic groups ($s=0$ and $s=1$) on a test set after training a model for 30 epochs. Panel~\ref{fig:DCF_optimization_0.0} shows the obtained score distributions for $\lambda=0$, i.e., a model optimized solely for classification error. If only instances in the decision-making region (i.e., those with a predicted score $\tilde{y} \geq \tau = 0.7$) are acted upon, this model would result in clearly unfair resource allocation. Panels~\ref{fig:DCF_optimization_0.3} and \ref{fig:DCF_optimization_0.6}, in contrast, induce decision-centric fairness with increasing strength, resulting in increasingly fair score distributions in the decision-making region (as reflected by the increasing overlap between the top-$k\%$ score distributions across demographic groups), potentially at the cost of predictive performance.

\section{Experimental design} \label{SEC:EXPERIMENTAL_DESIGN}

This section provides a detailed overview of the experimental design. 
We first describe the datasets used and the process of creating semi-synthetic data to introduce (additional) bias into the historical data,
providing a way to control the level of discriminatory behavior when left unmitigated.
This allows us to test the sensitivity of the different fairness induction methods to varying levels of bias in the data used to train classification models for resource allocation.
Next, we introduce and motivate the use of a decision-centric and threshold-independent measure based on the precision-recall curve to assess predictive performance, complementing the fairness evaluation using the decision-centric fairness metrics introduced in Section~\ref{SEC:METHODOLOGY_EVAL}.
Finally, we provide details on the problem configuration setups and the hyperparameter combinations tested.
The code for reproducing the experiments is publicly available at \url{https://github.com/SimonDeVos/DCF}.

\subsection{Data} \label{SEC:EXPERIMENTAL_DESIGN_DATA}
We use three datasets for our experiments: \textit{TelecomKaggle} (public), \textit{Churn} (proprietary), and \textit{Adult} (public).
The first two datasets are directly relevant in the context of resource allocation,
as they involve predicting churn propensity scores,
whereas the last one is included because it is a standard dataset in the algorithmic fairness literature\footnote{Note that it is difficult to find public datasets in the fields of credit risk management and fraud detection, as in these domains \textit{fairness through unawareness} is often used to comply with existing regulations; hence, no information on potential protected attributes is (publicly) available in those datasets \citep{chen2019fairness,coston2019fair}.}.
An overview of dataset characteristics is provided in Table~\ref{tab:datasets}.

The need for fairness induction arises when biases are present in the historical data used to train classification models, or when there are inherent differences between groups in their tendency to belong to the positive class, which---although statistically justified---are considered unacceptable discrimination when acted upon.
Specifically, if the data itself is unbiased and there are no inherent differences between groups,
a classification model optimized solely for classification error will naturally produce fair predicted scores,
and consequently, a fair resource allocation.
To test the sensitivity of the fairness induction methods discussed in Section~\ref{SEC:METHODOLOGY_TRAIN}, 
we control for the first issue mentioned above by varying the levels of discriminatory bias in the historical data for the \textit{TelecomKaggle} dataset.
Based on the raw \textit{TelecomKaggle} dataset, we create three semi-synthetic datasets to introduce (additional) bias into the historical data.
This bias is systematically introduced through \textit{informed label flipping}.
Specifically, within one protected group, we selectively flip ground-truth outcome labels from $0$ to $1$. 
The details of this procedure are outlined in Algorithm~\ref{alg:semisynth_data} in Appendix~\ref{app:dataset_details}.
Figures~\ref{fig:telecomkaggle_baseline_bias}--\ref{fig:adult_baseline_bias} display the score distributions obtained using no fairness induction, 
i.e., reflecting the baseline discriminatory behavior present in the datasets.
Figure~\ref{fig:telecomkaggle_baseline_bias} illustrates the effect of introducing additional bias through informed label flipping. As the bias increases, the violation of demographic parity becomes more pronounced---both globally across the entire domain $[0,1]$ and within the decision-making region $[\tau,1]$.

After introducing bias through informed label flipping (where applicable), each dataset is split into training, validation, and test sets using a fixed ratio of 0.34/0.33/0.33. The training set is used to learn model parameters, the validation set is used for hyperparameter tuning (i.e., selecting the configuration with the lowest validation loss for $\lambda=0$), and the test set is used for final model evaluation. A relatively large portion of the data is allocated to the validation and test sets to ensure sufficient resolution and stability in assessing performance and fairness metrics, particularly when analyzing decision-centric results across subgroups and thresholds.

\begin{table}[tb]
\centering
\begin{adjustbox}{max width=1.0\textwidth}
    \begin{tabular}{ccccc|cc|cc|cc}
        
    \toprule
        
        \multicolumn{1}{c}{\multirow{2}{*}{Dataset}} & \multicolumn{1}{c}{\multirow{1}{*}{Protected}} & \multirow{2}{*}{\# Vars.} & \multirow{2}{*}{\# Obs.} & \multirow{1}{*}{Bias} & \multicolumn{2}{c}{$s=0$} & \multicolumn{2}{c}{$s=1$} & \multicolumn{2}{l}{Class balance} \\
    
        \multicolumn{1}{l}{}                         &   attribute $s$&                             &                                  & rate                                      & $y=0$      & $y=1$      & $y=0$      & $y=1$      & $y=0$           & $y=1$           \\
    \midrule
    
        \multirow{3}{*}{TelecomKaggle}  & \multirow{3}{*}{`Sex'}               & \multirow{3}{*}{39}          & \multirow{3}{*}{7,032}                                  & 0.25                                  & 0.27       & 0.22       & 0.37       & 0.13       & 0.64            & 0.36            \\
         &                                            &                              &                                  & 0.50                                   & 0.18       & 0.31       & 0.37       & 0.13       & 0.55            & 0.45            \\
                                                     &                            &  &                                  & 0.75                                  & 0.09       & 0.40       & 0.37       & 0.13       & 0.46            & 0.54            \\
        \cdashline{1-11} 

        {Churn}  & `Sex'                             & 10                            & 44,942 & -                                     & 0.38       & 0.27       & 0.18       & 0.17       & 0.56            & 0.44           \\

                \cdashline{1-11} 

        {Adult}  &  `Sex'                              & 14                            & 32,561 & -                                     & 0.46       & 0.20      & 0.29       & 0.04        & 0.76            & 0.24           \\
    \bottomrule
    
    \end{tabular}
\end{adjustbox}
\caption{Overview of dataset characteristics. By introducing additional bias in \textit{TelecomKaggle}, 
we increase the ground-truth class imbalance within the protected group $s=1$.
For an overview of the score distributions obtained without fairness induction, i.e., reflecting the baseline discriminatory behavior present in the datasets, we refer to Figures~\ref{fig:telecomkaggle_baseline_bias}--\ref{fig:adult_baseline_bias} in Appendix~\ref{app:baseline_bias}.}
\label{tab:datasets}
\end{table}

\subsection{Evaluation metrics}\label{SEC:EXPERIMENTAL_DESIGN_EVALUATION_METRICS}

Our evaluation examines both predictive performance and fairness through a decision-centric lens,
enabling the assessment of how the deployment of the models would impact resource allocation.

\begin{figure}[tb]
    \centering
    \includegraphics[width=0.4\linewidth]{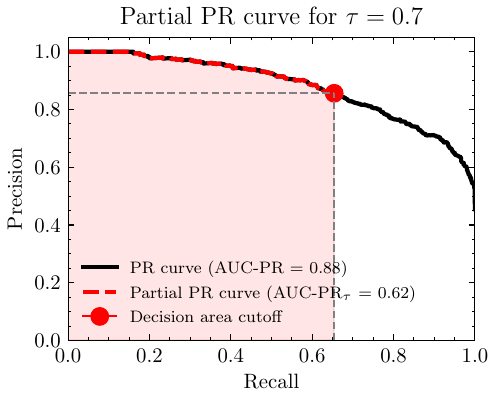}
    \caption{An example partial precision-recall curve and its corresponding $\text{AUC-PR}_\tau$, which summarizes precision and recall for all decision thresholds within the decision-making region $[\tau,1]$. This region corresponds to the top-left segment of the PR curve starting at $\tau$.}
    \label{fig:partial_aucpr}
\end{figure}

To evaluate the predictive performance of a classification model in the context of online resource allocation with a dynamic decision threshold $\tau$,
driven by dynamic resource constraints (and/or instance-dependent costs and benefits),
we need a threshold-independent evaluation metric.
Moreover, this metric should focus on predictions within the decision-making region, as these are the predictions we will potentially act upon. Specifically, for predictions across the relevant decision thresholds $[\tau, 1]$, we aim to maximize both precision and recall.
Focusing on precision alone can be misleading, as a model with fewer predictions for which $\tilde{y} \geq \tau$ may appear more precise simply because it predicts $\tilde{y} \geq \tau$ only for the most confident cases, disregarding recall. Conversely, focusing solely on recall may reward models that capture more true positives at the cost of a higher false positive rate, leading to wasted resources.
Precision-recall (PR) curves address this trade-off by considering both metrics simultaneously for all possible thresholds.
To restrict the thresholds of interest to those within the decision-making region, we introduce the decision-centric 
performance metric $\text{AUC-PR}_{\tau}$.
This metric summarizes precision and recall by calculating the area under a partial PR curve, which only considers thresholds above $\tau$.
This ensures that the performance assessment incorporates the deployment of the classification model in combination with all decision thresholds within the decision-making region.
Figure~\ref{fig:partial_aucpr} illustrates how $\text{AUC-PR}_{\tau}$ is obtained.

In addition to evaluating decision-centric predictive performance via $\text{AUC-PR}_{\tau}$, we also evaluate the decision-centric fairness of the classification models.
To this end, we use the $\text{ABPC}_{\tau}$ 
(Equation~\eqref{eq:abpc_local}) and $\text{ABCC}_{\tau}$ (Equation~\eqref{eq:abcc_local}) metrics introduced in Section~\ref{SEC:METHODOLOGY_EVAL}.

\subsection{Problem and hyperparameter configurations}

The most important hyperparameter in our experimental study is $\lambda$,
as it controls the importance assigned to the unfairness penalty during training.
Due to the predictive performance-fairness trade-off (see Section~\ref{SEC:DP_IN_RESOURCE_ALLOCATION}),
however, objectively tuning this hyperparameter is often not possible.
Therefore, we train neural classification models for $\lambda \in \{0,0.05,0.10,\dots,0.95\}$ for both the global and decision-centric fairness approaches.
These models will be assessed using the concept of Pareto-optimality, as explained in more detail in Section~\ref{SEC:RESULTS+DISCUSSION}.
To determine all other hyperparameters, tuning is performed for the model with $\lambda = 0$ on each dataset.
The hyperparameter configuration that results in the lowest validation loss is selected, and the same configuration is subsequently used for all values of $\lambda$.
To quantify dissimilarities between score distributions for different demographic groups,
i.e., for $\mathcal{L}_\text{unfairness}$,
we use the implementation by \citet{shalit2017estimating}
to approximate the 1-Wasserstein distance IPM \citep{villani2008optimal} (and its gradients) using Sinkhorn distances \citep{cuturi2013sinkhorn,cuturi2014wasserstein}.
For further details, we refer to Appendix B.1 of \citet{shalit2017estimating}.
Additional information on the model architecture, implementation, training, and hyperparameter tuning is provided in Appendix~\ref{app:implementation}.

In addition to testing the sensitivity of the different fairness induction methods to varying levels of discriminatory bias in the data
(via the creation of semi-synthetic datasets with additional bias introduced; see Section~\ref{SEC:EXPERIMENTAL_DESIGN_DATA}),
we also assess their sensitivity to the size of the decision-making region by varying the decision threshold $\tau$.

\section{Results and discussion} \label{SEC:RESULTS+DISCUSSION}
In this section, we present our results by comparing the Pareto fronts of the global and decision-centric fairness induction approaches.
Specifically, we visualize the trade-offs achieved between decision-centric predictive performance---measured by $\text{AUC-PR}_{\tau}$ (higher is better)---and decision-centric fairness---measured by $\text{ABPC}_{\tau}$ or $\text{ABCC}_{\tau}$ (lower is better)---for the different values of $\lambda$; and we construct Pareto fronts for both the global and decision-centric approaches by identifying all models that represent optimal trade-offs under Pareto-optimality (i.e., models for which no objective can be improved without worsening the other).
The model without fairness induction (i.e., $\lambda = 0$) is also visualized and marked with a red star.

We structure our presentation and discussion of the results around the following three key questions:
\begin{itemize}
    \item \textbf{Q1}: What is the impact of adopting a decision-centric versus a global approach to inducing fairness in the decision-making region on predictive performance?
    \item \textbf{Q2}: How do the size of the decision-making region and varying levels of discriminatory bias in the historical data affect the differences in predictive performance between a decision-centric versus a global fairness induction approach?
    \item \textbf{Q3}: Which metric is most appropriate for evaluating decision-centric fairness, and how should a model be selected for deployment?
\end{itemize}
To address these questions, we selectively highlight relevant results. A comprehensive set of results is presented in Appendix~\ref{app:extra_results}.

\subsection{Q1: Impact of decision-centric versus global fairness approach on predictive performance}

\begin{figure}[t!]
\centering
\begin{subfigure}[b]{0.3\textwidth}
    \includegraphics[width=\textwidth]{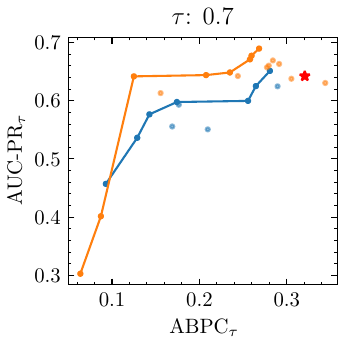}
    \caption{\textit{TelecomKaggle} (bias rate 0.5)}
    \label{fig:telecomkaggle_abpc_bias0.50_pcta0.7}
\end{subfigure}
\hspace{0.03\textwidth}
\begin{subfigure}[b]{0.3\textwidth}
    \includegraphics[width=\textwidth]{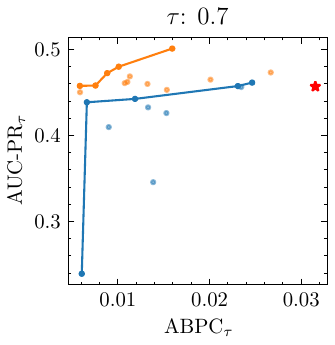}
    \caption{\textit{Churn}}
    \label{fig:korea6_abpc_pcta0.7}
\end{subfigure}
\hspace{0.03\textwidth}
\begin{subfigure}[b]{0.3\textwidth}
    \includegraphics[width=\textwidth]{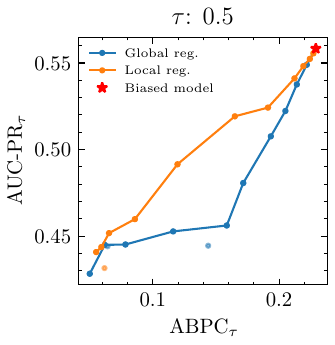}
    \caption{\textit{Adult}}
    \label{fig:adult_abpc_pcta0.5}
\end{subfigure}
\caption{
Pareto fronts illustrating the trade-off between predictive performance and fairness ($\text{AUC-PR}_{\tau}$ vs.\ $\text{ABPC}_{\tau}$) for the decision-centric (orange) and global (blue) fairness induction approaches for three datasets.
}
\label{fig:results_q1_abpc}
\end{figure}

\begin{figure}[t!]
\centering
\begin{subfigure}[b]{0.3\textwidth}
    \includegraphics[width=\textwidth]{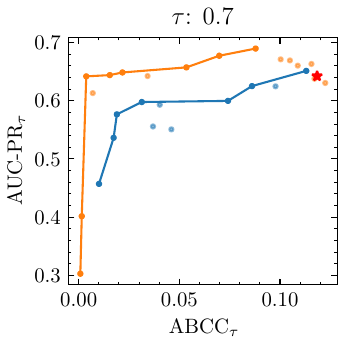}
    \caption{\textit{TelecomKaggle} (bias rate 0.5)}
    \label{fig:telecomkaggle_abcc_bias0.50_pcta0.7}
\end{subfigure}
\hspace{0.03\textwidth}
\begin{subfigure}[b]{0.3\textwidth}
    \includegraphics[width=\textwidth]{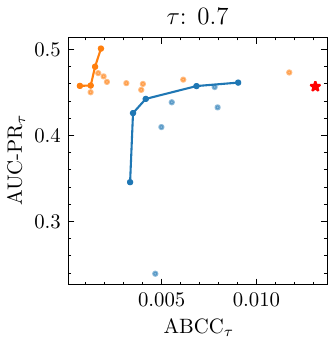}
    \caption{\textit{Churn}}
    \label{fig:korea6_abcc_pcta0.7}
\end{subfigure}
\hspace{0.03\textwidth}
\begin{subfigure}[b]{0.3\textwidth}
    \includegraphics[width=\textwidth]{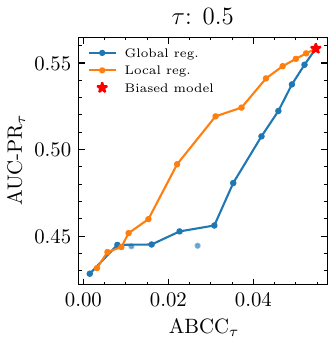}
    \caption{\textit{Adult}}
    \label{fig:adult_abcc_pcta0.5}
\end{subfigure}
\caption{
Pareto fronts illustrating the trade-off between predictive performance and fairness ($\text{AUC-PR}_{\tau}$ vs.\ $\text{ABCC}_{\tau}$) for the decision-centric (orange) and global (blue) fairness induction approaches for three datasets.
}
\label{fig:results_q1_abcc}
\end{figure}

We first investigate how decision-centric fairness optimization compares to the global approach by evaluating models trained with varying values of $\lambda$. Figures~\ref{fig:results_q1_abpc} and \ref{fig:results_q1_abcc} show the results for three datasets using $\text{ABPC}_{\tau}$ and $\text{ABCC}_{\tau}$ as the fairness metric, respectively.

The plots clearly illustrate the advantage of our proposed decision-centric method (orange curves) over the global approach (blue curves). As $\lambda$ increases, model performance moves along a trade-off curve, prioritizing fairness at the expense of predictive performance. However, the decision-centric approach consistently delivers superior trade-offs, underscoring its effectiveness in balancing fairness with predictive performance specifically within the decision-making region.

The trade-off varies by dataset. For example, in the \textit{TelecomKaggle} and \textit{Churn} datasets, fairness can be substantially improved with relatively minor reductions in predictive performance. 
Conversely, for the \textit{Adult} dataset, even a slight increase in $\lambda$ leads to a drop in predictive performance, though the decision-centric approach still yields a more favorable trade-off than the global method.

Moreover, adding an unfairness penalty---whether decision-centric or global---can further enhance both fairness and predictive performance. This effect is visible in the plots, where certain models lie higher and further to the left than the model without unfairness penalty (i.e., with $\lambda=0$) marked with a red star. This shows that the unfairness penalty, in some cases, also serves as a regularization mechanism.

\subsection{Q2: Impact of decision-making region size and level of discriminatory bias in historical data}

\begin{figure}[t!]
    \centering
    \includegraphics[width=0.8\linewidth]{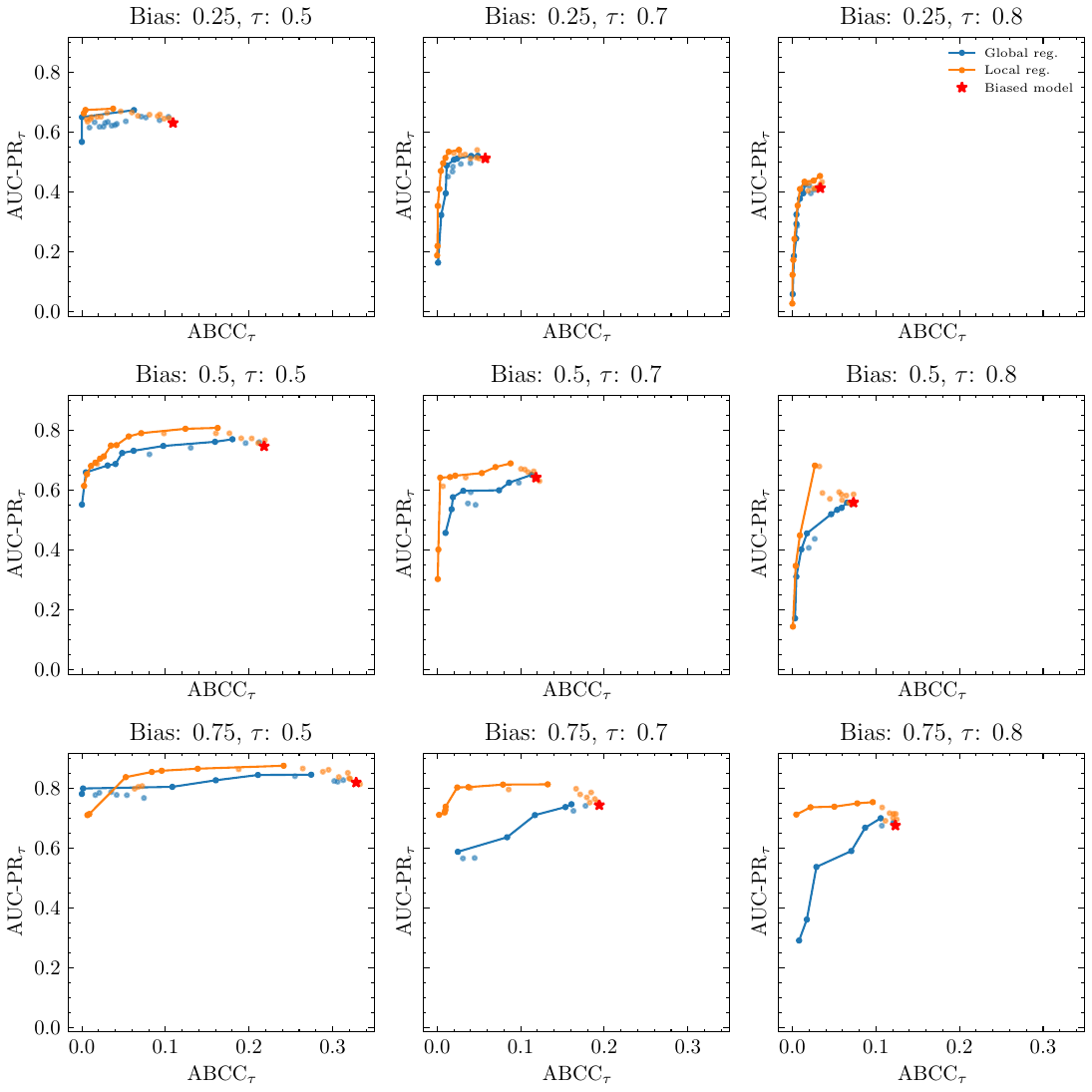}
    \caption{    
    Results on the \textit{TelecomKaggle} dataset for different bias rates, with decision-centric fairness measured by $\text{ABCC}_{\tau}$.  
    The figure illustrates the effect of a varying size of the decision-making region with $\tau=0.5, 0.7, 0.8$ (columns) on the decision-centric fairness-predictive performance trade-off.  
    The orange and blue lines represent decision-centric and global fairness induction, respectively, while the model without unfairness penalty (i.e., with $\lambda=0$) is marked with a red star.}
    \label{fig:results_telecomkaggle_abcc}
\end{figure}

We analyze the influence of varying the decision threshold $\tau$ and semi-synthetic bias rates on model performance, both in terms of predictive performance and fairness, using the \textit{TelecomKaggle} dataset.
Using the informed label flipping method to generate semi-synthetic datasets with varying bias levels allows us to systematically investigate the impact of bias on model outcomes. Results are shown in Figure~\ref{fig:results_telecomkaggle_abcc}, where rows represent different bias rates (0.25, 0.50, 0.75) and columns represent different decision thresholds ($\tau$ values of 0.5, 0.7, and 0.8), with fairness measured by $\text{ABCC}_{\tau}$.
For fairness results in terms of $\text{ABPC}_{\tau}$, see Figure~\ref{fig:results_telecomkaggle_abpc} in the appendix.

The models without unfairness penalization (i.e., with $\lambda=0$, indicated by red stars) illustrate how fairness deteriorates as either the bias increases or the decision-making region becomes larger.
Specifically, for a fixed $\tau$, higher semi-synthetic bias consistently reduces fairness, evidenced by a rightward shift of baseline points. 
Similarly, for a fixed bias rate, a larger decision-making region (lower $\tau$) also negatively impacts fairness. 
This is because more predictions are included in the decision-making region and thus affect decision-centric fairness calculations. 
Consequently, the best baseline decision-centric fairness is observed with minimal bias and a higher $\tau$, while the worst-case scenario arises from a combination of high bias and a large decision-making region.

In comparing global (blue) and decision-centric (orange) approaches, we observe similar performance for the low bias rate (0.25), where fairness induction has a minimal potential impact as there is little bias present to be eliminated. 
However, for the higher bias rates (0.50 and 0.75), the decision-centric approach increasingly outperforms the global approach, demonstrating its effectiveness.
The advantage of the decision-centric approach over the global approach becomes more pronounced as the decision threshold $\tau$ increases, resulting in a smaller decision-making region.
Conversely, in the extreme case where $\tau=0$, both decision-centric and global fairness induction methods coincide due to the decision-making region covering the entire prediction domain.

For results on varying the decision threshold $\tau$ for the \textit{Churn} and \textit{Adult} datasets, we refer to Figures~\ref{fig:results_koreachurn_abpc}--\ref{fig:results_adult_abcc} in the appendix. For the \textit{Adult} dataset, 
in addition to $\tau$ values of $0.5$ and $0.8$,
we also include $\tau=0.4$ because of the class imbalance, which leads to fewer predictions in the decision-making region.

\subsection{Q3: Impact of decision-centric fairness metric used for evaluation and model selection}

The ABCC metric directly evaluates the 1-Wasserstein distance between the score distributions of different demographic groups \citep{han2023}, and as such, it aligns closely with our implementation of $\mathcal{L}_\text{unfairness}$ for the global approach.
A similar alignment exists between the $\text{ABCC}_\tau$ metric and the decision-centric fairness approach,
which may (partly) explain the larger differences observed between the Pareto fronts of the global and decision-centric approaches for the \textit{TelecomKaggle} and \textit{Churn} datasets (compare the plots in Figures~\ref{fig:results_q1_abpc} and \ref{fig:results_q1_abcc}).

Beyond the optimization-evaluation alignment, while the PDF-based $\text{ABPC}_\tau$ metric provides an intuitive way to quantify decision-centric fairness (see Figure~\ref{fig:intro_toy}), the $\text{ABCC}_\tau$ metric is also better suited to the underlying problem of using predicted scores for resource allocation,
as it is sensitive to the distance that probability mass must move to align the score distributions across groups---capturing not only the existence but also the magnitude of score shifts.
In contrast, ABPC (and therefore $\text{ABPC}_\tau$) only quantifies how much probability mass is placed differently between groups, but is invariant to how far these differences are from each other.
For instance, if two distributions differ only in localized regions---say, around 0.75 and 0.9 in one scenario, versus around 0.85 and 0.9 in another (with the density being lower around one value and higher around the other)---then $\text{ABPC}_\tau$ with $\tau < 0.7$ would yield the same value for both. However, $\text{ABCC}_\tau$ would favor the latter scenario.
Since the likelihood of these differences---and thus the algorithmic unfairness---being canceled out through thresholding (with the decision threshold varying between $\tau$ and 1) increases when the distances between them are smaller, $\text{ABCC}_\tau$ is preferable for evaluating decision-centric fairness in resource allocation contexts.    

Hence, to select a classification model for deployment in the context of resource allocation optimization, we advise decision-makers to choose a model (which, in our experiments, corresponds to selecting a fairness induction strategy together with a value for $\lambda$) that offers a good trade-off between $\text{ABCC}_\tau$ and $\text{AUC-PR}_{\tau}$.
Our results show that it is often possible to substantially reduce $\text{ABCC}_\tau$ through decision-centric fairness induction (and in some cases also through global fairness induction), without compromising $\text{AUC-PR}_{\tau}$ compared to a model without fairness induction.
As such, when fairness is not mandated by strict regulatory requirements, these models are of particular interest.    
Performance in terms of $\text{ABPC}_\tau$ can be used to further assess a (small) subset of candidate models in a complementary manner.

\section{Conclusion} \label{SEC:CONCLUSION}

This paper introduces and formalizes the concept of decision-centric fairness in classification models used for online resource allocation optimization with dynamic resource constraints. This novel approach aligns algorithmic fairness considerations with their potential impact on real-world resource allocation decisions, such as the fairness of credit risk scores used for loan approvals or churn propensity scores used in targeted marketing retention campaigns, while accommodating dynamic decision thresholds.
Decision-centric fairness redefines fairness evaluation and induction by shifting the focus from full classification score distributions to considering only the decision-making region, which includes only the range of relevant decision thresholds for a given resource allocation problem. This ensures that fairness constraints are considered and/or enforced only where they will impact real-world decisions.
We propose a method to optimize classification models directly for decision-centric fairness and demonstrate that by avoiding overly strict fairness constraints, we can minimize the degradation of the predictive quality of the generated scores.
Specifically, we empirically show that our decision-centric fairness methodology often leads to better decision-centric predictive performance-fairness trade-offs compared to a global fairness approach, which applies fairness constraints more naively across the full score distributions.

For certain resource allocation problems, such as targeted retention campaigns based on churn propensity scores, a causal uplift modeling approach has been shown to outperform the predictive modeling approach adopted in this work, leading to improved profitability of retention campaigns \citep{devriendt2021uplift}.
However, the predictive modeling approach remains valuable in practice, as it allows the use of a single customer churn prediction model across various retention campaigns---something that is not possible with the treatment-dependent uplift modeling approach.
Nevertheless, adapting our proposed decision-centric fairness approach to optimize uplift scores appears to be a promising direction for future research.
Exploring this in the context of cost-sensitive uplift modeling \citep{verbeke2023cscc}, which leverages class-dependent or instance-dependent costs and benefits to improve decision-making, also seems worthwhile, though potentially more challenging. As a first step, incorporating decision-centric fairness into the cost-sensitive predictive modeling approach \citep{verbeke2012churn} could serve as a starting point.
In a similar vein to \citet{devriendt2021uplift}, \citet{vanderschueren2024l2r} recently demonstrated that framing resource allocation problems with stochastic resource constraints as a ranking problem, and subsequently relying on learning-to-rank techniques instead of classification techniques,
leads to improved decision-making outcomes.
This ranking approach naturally applies to all resource allocation problems (in contrast to an uplift modeling approach).
Hence, incorporating a decision-centric fairness perspective into the optimization of learning-to-rank models appears to be a promising direction for future research as well.
Translating such fairness-enhanced learning-to-rank models to the uplift modeling setting,
building on \citet{devriendt2020l2r}, would help complete the picture.


Beyond these methodological extensions, our work opens several additional avenues for future research concerning the fairness perspective.
First, the fairness metrics used in this study center on a single group fairness notion: decision-centric demographic parity. 
While practical and grounded in regulatory logic \citep{EU2012Guidelines, feldman2015certifying}, group-based metrics have notable limitations as they mask disparities at the individual level. 
Moreover, there might be adverse effects of reranking individuals (across or within sensitive groups), which may undermine the perceived validity of fairness \citep{goethals2024reranking}.
Future research could examine alternative or complementary notions of fairness, such as equal opportunity \citep{hardt2016} or counterfactual fairness \citep{kusner_2017_counterfactuals}. 
Second, we adopt a percentile-based approach to operationalize decision-centric fairness induction---aligning the top$-k\%$ of each group’s score distribution.
This approach, however, remains a proxy for the ideal case where fairness is enforced strictly within the decision-making region. 
As discussed in Section~\ref{SEC:METHODOLOGY_TRAIN}, a quantile-based method that directly compares and penalizes disparities in score distributions above a threshold $\tau$ would align more closely with the formalization of decision-centric fairness.
However, our initial experiments with such quantile-based implementations revealed training instability. Therefore, investigating how to efficiently and robustly implement such a quantile-based approach is left for future research.
Next, although not yet operationalized, decision-centric fairness could naturally extend to multiple or intersecting protected attributes, suggesting a direction for future work \citep{barocas_fairml_2019,binns_apparent_2020}.
Finally, further strengthening the formal connection between decision-centric fairness and legal interpretations of \textit{actionability} in anti-discrimination law \citep{feldman2015certifying} would be a valuable step toward ensuring legal as well as practical soundness. 
Bridging the conceptual gap between legal standards of fairness and the integration of fairness considerations into algorithm evaluation and design remains a critical step in operationalizing fairness.

\section*{Acknowledgements}
Simon De Vos is supported by Acerta Consult and the Flemish Government through Flanders Innovation \& Entrepreneurship (VLAIO) [project HBC.2021.0833].
Jente Van Belle and Andres Algaba gratefully acknowledge financial support from the Research Foundation – Flanders [grant numbers 12AZX24N and 1286924N].





\newpage
\appendix

\renewcommand{\thesection}{\Alph{section}}
\setcounter{table}{0}
\renewcommand{\thetable}{\thesection\arabic{table}}
\setcounter{figure}{0}
\renewcommand{\thefigure}{\thesection\arabic{figure}}

\section{Dataset details}\label{app:dataset_details}

\subsection{Label flipping for inducing additional bias}

\begin{algorithm}[]
\floatname{algorithm}{Algorithm}
\caption{Informed label flipping for semi-synthetic dataset generation}\label{alg:semisynth_data}

\textbf{Input:} $\mathcal{D}=\{(\bm{x}_i,y_i,s_i)\}_{i=1}^N$, 
protected group value $s \in \{0,1\}$,
bias rate $r \in [0,1]$ \\
\textbf{Output:} Biased dataset ${\mathcal{D}'_r}$

\begin{algorithmic}[1]
    \State{Train a classifier $m$ on $\mathcal{D}$}
    \State $\tilde{y} \gets m(\bm{x},s)$ \Comment{Predicted scores $\tilde{y}_i \in [0,1]$}
    \State $\mathcal{D}^s_0 \gets \{(\bm{x}_i, y_i, s_i, \tilde{y}_i) \mid s_i = s \text{ and } y_i = 0\}$       \Comment{Subset for bias introduction}
    \State $k \gets r \cdot |\mathcal{D}^s_0|$
    \State Rank $\mathcal{D}^s_0$ by $\tilde{y}$
    \For {$(\bm{x}_i, y_i) \in \text{top-}k \text{ of } \mathcal{D}^s_0$}     
        \State ${y}_i \gets 1$            \Comment{Flip labels}
    \EndFor
    \State \Return ${\mathcal{D}'_r} = \{(\bm{x}_i,y_i,s_i)\}_{i=1}^N$   \Comment{Return biased dataset}
\end{algorithmic}
\end{algorithm}

\subsection{Baseline discriminatory behavior}\label{app:baseline_bias}

\begin{figure}[h]
\centering
    \begin{subfigure}[b]{0.30\textwidth}
    \includegraphics[width=\textwidth]{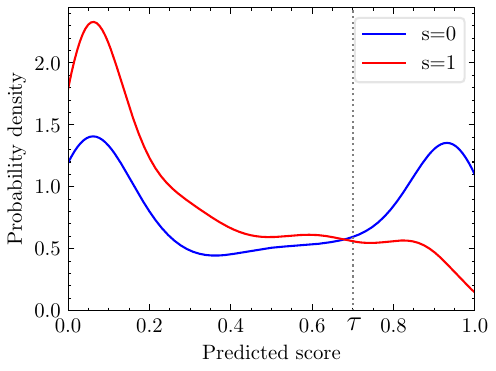}
    \caption{Bias rate $0.25$}
    \label{fig:bias25}
\end{subfigure}
\hfill
\begin{subfigure}[b]{0.30\textwidth}
    \includegraphics[width=\textwidth]{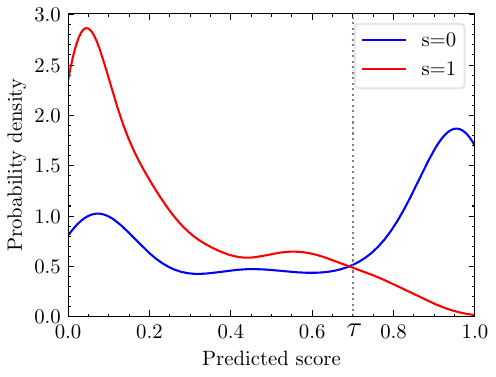}
    \caption{Bias rate $0.50$}
    \label{fig:bias50}
\end{subfigure}
\hfill
\begin{subfigure}[b]{0.30\textwidth}
    \includegraphics[width=\textwidth]{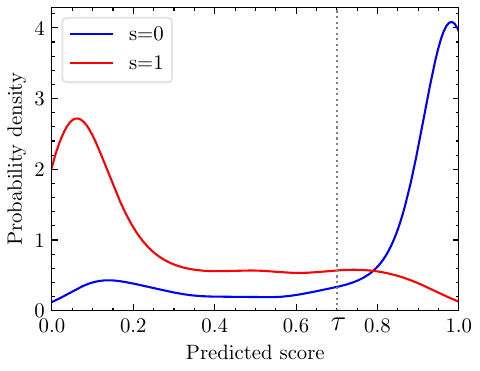}
    \caption{Bias rate $0.75$}
    \label{fig:bias75}
\end{subfigure}
\caption{
Baseline discriminatory behavior for the \textit{TelecomKaggle} datasets.
This figure shows the score distributions (PDFs) for two groups ($s=0$ and $s=1$), as estimated by a model without unfairness penalty ($\lambda=0$), under three semi-synthetic bias rates. 
The decision-making region is defined as $\tau=0.7$.
Higher semi-synthetic bias rates lead to increased baseline discriminatory behavior, as summarized below in Table~\ref{tab:baseline_bias}.
}\label{fig:telecomkaggle_baseline_bias}
\end{figure}

\begin{figure}[h]
    \begin{minipage}{0.45\linewidth}
        \centering
        \includegraphics[width=\linewidth]{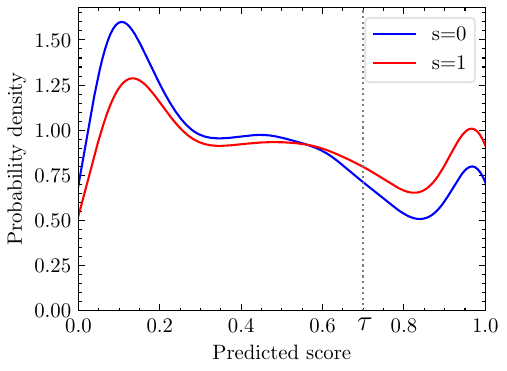}
        \caption{Baseline discriminatory behavior for the \textit{Churn} dataset.}
        \label{fig:korea6_baseline_bias}
    \end{minipage}
    \hfill
    \begin{minipage}{0.45\linewidth}
        \centering
        \includegraphics[width=\linewidth]{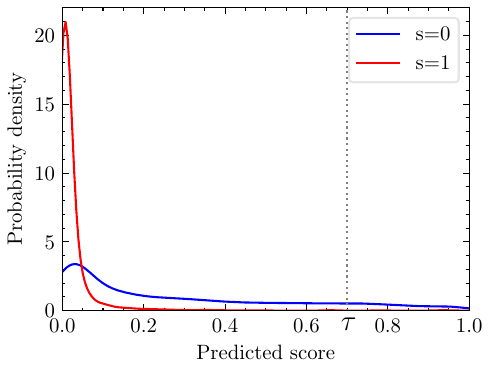}
        \caption{Baseline discriminatory behavior for the \textit{Adult} dataset.}
        \label{fig:adult_baseline_bias}
    \end{minipage}
\end{figure}

\begin{table}[h]
\centering
\begin{tabular}{lc|cc}
\toprule
Dataset & Bias rate & {$\text{ABPC}_{\tau}$} & {$\text{ABCC}_{\tau}$} \\
\midrule
\multirow{3}{*}{\textit{TelecomKaggle}}     & 0.25           & 0.16 & 0.06 \\
                                            & 0.50           & 0.30 & 0.12 \\
                                            & 0.75           & 0.43 & 0.19 \\

\cdashline{1-4}

\textit{Churn}                         & - & 0.03 & 0.01 \\

\cdashline{1-4}

\textit{Adult}                              & - & 0.13 & 0.02 \\

\bottomrule
\end{tabular}
\caption{Baseline discriminatory behavior for the various datasets for $\tau=0.7$, corresponding to Figures~\ref{fig:telecomkaggle_baseline_bias}-\ref{fig:adult_baseline_bias}.}
\label{tab:baseline_bias}
\end{table}

\section{Implementation}\label{app:implementation}

We employ a fully-connected MLP, implemented in PyTorch, where network parameters are optimized using Adam optimizer \citep{kingma2014adam} with a learning rate $0.01$ for \textit{TelecomKaggle} and \textit{Churn}, and $0.001$ for \textit{Adult}.
We use ReLu activation functions in the hidden layers. 
For the final layer, we use a sigmoid activation function to ensure outputs in the range $[0,1]$. 
We use a standard BCE loss function for the first 15 epochs and then a composite loss function for the remainder of the training.
We employ early stopping after 20 epochs without model improvement in terms of composite loss on the validation data.
Per dataset, hyperparameters are selected for the case where $\lambda=0$.
To ensure enough observations remain for computing the decision-centric fairness loss---given that each training batch considers only a subset of data restricted to the top $n\%$ percentile, further divided across the two protected groups and limited to the training split---we set the batch size to $1024$. 

\begin{table}[h]
\centering
\begin{tabular}{ll ccccc}
\toprule
\textit{Hyperparameter} & {Values} & \multicolumn{3}{c}{{TelecomKaggle}} & \multirow{1}{*}{{Churn}} & \multirow{1}{*}{{Adult}} \\
\cmidrule(lr){3-5}

                                &                       &  0.25 & 0.50 & 0.75 &  &\\

\midrule

\textit{Nr.\ of hidden layers}   & \{2, 3, 4\}           & 3     & 2     & 4     & 2     & 2     \\
\textit{Hidden layer size }     & \{16, 32, 64, 128\}   & 64    & 64    & 32    & 32    & 64    \\
\textit{Dropout probability}    & \{0, 0.01, 0.1\}      & 0     & 0     & 0     & 0.01  & 0.01  \\
\textit{L2 regularization }     & \{0, 0.01, 0.05\}     & 0.01  & 0.01  & 0.01  & 0     & 0.01  \\

\bottomrule
    \end{tabular}
    \caption{Hyperparameter search space. }
    \label{tab:hyperparameters}
\end{table}

\clearpage
\section{Additional results}\label{app:extra_results}

\begin{figure}[h]
    \centering
    \includegraphics[width=0.8\linewidth]{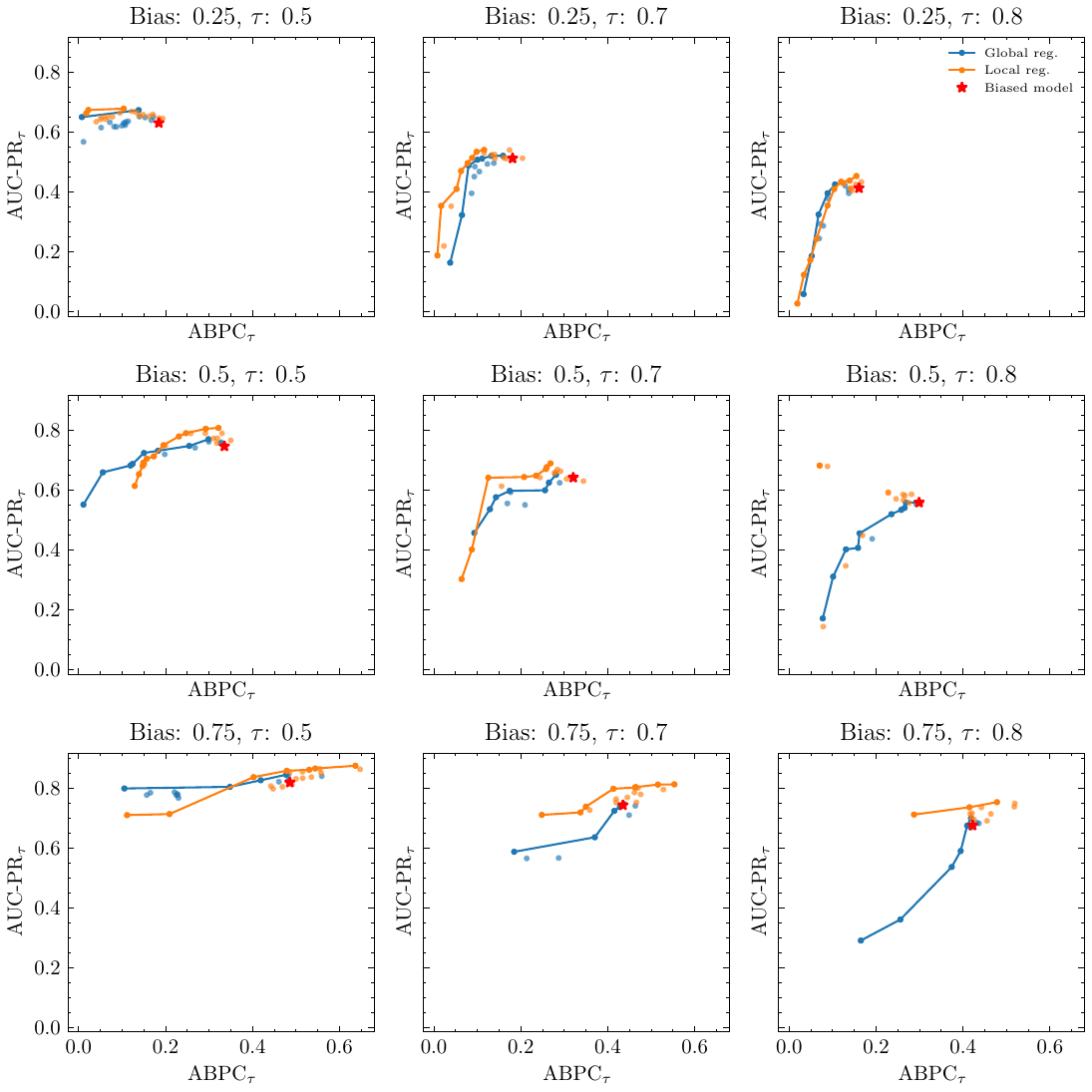}
\caption{
    Results on the \textit{TelecomKaggle} dataset for different bias rates, with fairness measured by $\text{ABPC}_{\tau}$.  
    The figure illustrates the effect of a varying size of the decision-making region with $\tau=0.5, 0.7, 0.8$ (columns) on the decision-centric fairness-predictive performance trade-off.  
    The orange and blue lines represent decision-centric and global fairness induction, respectively, while the model without unfairness penalty (i.e., with $\lambda=0$) is marked with a red star.  
}
    \label{fig:results_telecomkaggle_abpc}
\end{figure}

\begin{figure}[h]
    \centering
    \includegraphics[width=0.8\linewidth]{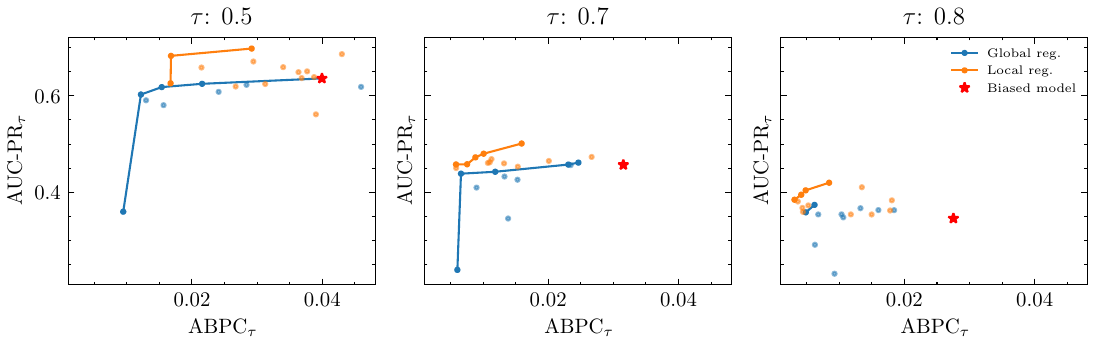}
    \caption{
    Results on the \textit{Churn} dataset with fairness measured by $\text{ABPC}_{\tau}$.
    The figure illustrates the effect of a varying size of the decision-making region with $\tau=0.5, 0.7, 0.8$ (columns) on the decision-centric fairness-predictive performance trade-off.  
    The orange and blue lines represent decision-centric and global fairness induction, respectively, while the model without unfairness penalty (i.e., with $\lambda=0$) is marked with a red star.  
    }
    \label{fig:results_koreachurn_abpc}
\end{figure}

\begin{figure}[h]
    \centering
    \includegraphics[width=0.8\linewidth]{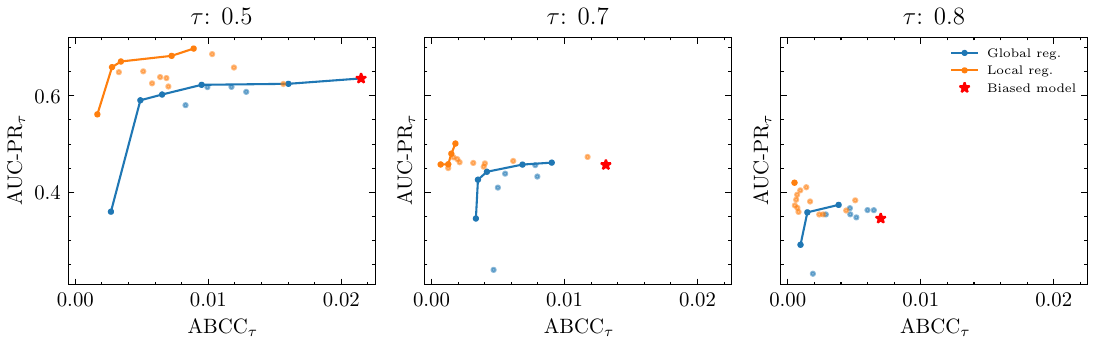}
    \caption{    
    Results on the \textit{Churn} dataset with fairness measured by $\text{ABCC}_{\tau}$.
    The figure illustrates the effect of a varying size of the decision-making region with $\tau=0.5, 0.7, 0.8$ (columns) on the decision-centric fairness-predictive performance trade-off.  
    The orange and blue lines represent decision-centric and global fairness induction, respectively, while the model without unfairness penalty (i.e., with $\lambda=0$) is marked with a red star.  
    }
    \label{fig:results_koreachurn_abcc}
\end{figure}

\begin{figure}[h]
    \centering
    \includegraphics[width=1\linewidth]{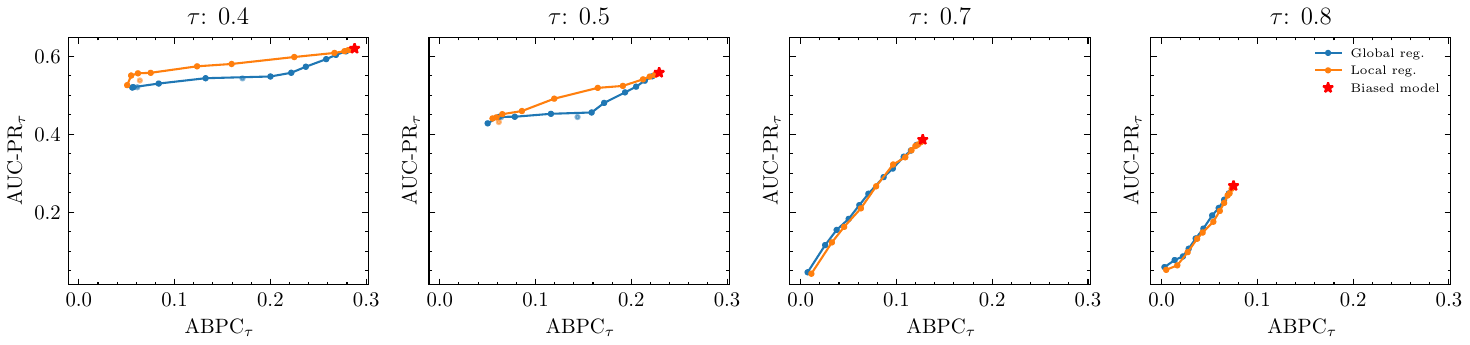}
    \caption{
    Results on the \textit{Adult} dataset with fairness measured by $\text{ABPC}_{\tau}$.
    The figure illustrates the effect of a varying size of the decision-making region with $\tau=0.4, 0.5, 0.7, 0.8$ (columns) on the decision-centric fairness-predictive performance trade-off.  
    The orange and blue lines represent decision-centric and global fairness induction, respectively, while the model without unfairness penalty (i.e., with $\lambda=0$) is marked with a red star.  
    }
    \label{fig:results_adult_abpc}
\end{figure}

\begin{figure}[h]
    \centering
    \includegraphics[width=1\linewidth]{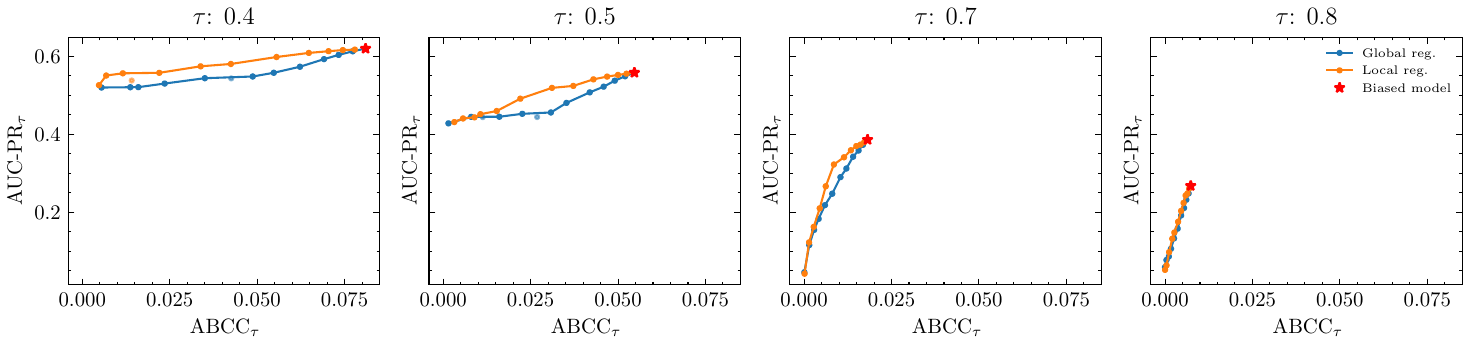}
    \caption{    
    Results on the \textit{Adult} dataset with fairness measured by $\text{ABCC}_{\tau}$.
    The figure illustrates the effect of a varying size of the decision-making region with $\tau=0.4, 0.5, 0.7, 0.8$ (columns) on the decision-centric fairness-predictive performance trade-off.  
    The orange and blue lines represent decision-centric and global fairness induction, respectively, while the model without unfairness penalty (i.e., with $\lambda=0$) is marked with a red star.  
    }
    \label{fig:results_adult_abcc}
\end{figure}

\clearpage

\bibliographystyle{elsarticle-harv} 
\bibliography{bibliography}


\end{document}